\documentclass[10pt,letterpaper]{article}
\usepackage[utf8]{inputenc}
\usepackage{float}

\usepackage{ccn}
\usepackage{graphicx}
\usepackage{pslatex}
\usepackage{apacite}
\usepackage{hyperref}
\usepackage{natbib}
\usepackage{lineno}
\usepackage{amsmath} 
\usepackage{xcolor}

\usepackage[normalem]{ulem}
\usepackage{gensymb}

\title{Higher-Order Convolution Improves Neural Predictivity in the Retina}

\author{\large \bf Simone Azeglio (simone.azeglio@gmail.com) \\ 
Institut de la Vision \& Laboratoire des Systèmes Perceptifs \\
Sorbonne Université, CNRS, INSERM, \& École Normale Supérieure\\
17 rue Moreau, 75012, Paris, France \& 24 rue Lhomond, 75005, Paris, France \\
\AND
\large \bf Victor Calbiague Garcia, Guilhem Glaziou \\ 
Institut de la Vision\\ 
Sorbonne Université, CNRS, INSERM, \\
17 rue Moreau, 75012, Paris, France\\
\AND 
\large \bf  Peter Neri\\ 
Sensory Processing and Computation Unit \& Laboratoire des Systèmes Perceptifs\\
Italian Institute of Technology \& École Normale Supérieure \\
Via Enrico Melen, 83, 16152 Genova, Italy \&  24 rue Lhomond, 75005, Paris, France\\
\AND
\large \bf Olivier Marre, Ulisse Ferrari \\ 
Institut de la Vision\\ 
Sorbonne Université, CNRS, INSERM, \\
17 rue Moreau, 75012, Paris, France\\
}

\begin{document}

\maketitle

\section{Abstract}
{
\bf
We present a novel approach to neural response prediction that incorporates higher-order operations directly within convolutional neural networks (CNNs).
Our model extends traditional 3D CNNs by embedding higher-order operations within the convolutional operator itself, enabling direct modeling of multiplicative interactions between neighboring pixels across space and time. Our model increases the representational power of CNNs without increasing their depth, therefore addressing the architectural disparity between deep artificial networks and the relatively shallow processing hierarchy of biological visual systems.
We evaluate our approach on two distinct datasets: salamander retinal ganglion cell (RGC) responses to natural scenes, and a new dataset of mouse RGC responses to controlled geometric transformations. Our higher-order CNN (HoCNN) achieves superior performance while requiring only half the training data compared to standard architectures, demonstrating correlation coefficients up to 0.75 with neural responses (against 0.80$\pm$0.02 retinal reliability). When integrated into state-of-the-art architectures, our approach consistently improves performance across different species and stimulus conditions.
Analysis of the learned representations reveals that our network naturally encodes fundamental geometric transformations, particularly scaling parameters that characterize object expansion and contraction. 
This capability is especially relevant for specific cell types, such as transient OFF-alpha and transient ON cells, which are known to detect looming objects and object motion respectively, and where our model shows marked improvement in response prediction. The correlation coefficients for scaling parameters are more than twice as high in HoCNN (0.72) compared to baseline models (0.32).
Our results demonstrate that incorporating biologically-inspired computational primitives can lead to more efficient visual processing without increasing architectural depth. This approach not only improves neural response prediction, but also provides insights into how biological systems might implement sophisticated computations within relatively few processing stages.
}

\begin{quote}
\small
\textbf{Keywords:} 
Higher-order convolution, Neural Response Prediction, Retinal Ganglion Cells, Neural Representations, Visual Processing, Alignment, Mechanistic Interpretability.
\end{quote}

\section{Introduction}

Convolutional Neural Networks (CNNs) have become fundamental tools for predicting neural responses in visual neuroscience \citep{yamins2016using, kriegeskorte2015deep, lindsay2021convolutional}. However, current state-of-the-art models, such as those evaluated on Brain-Score \citep{SchrimpfKubilius2018BrainScore, Schrimpf2020integrative} or Sensorium \citep{willeke2022sensorium, willeke2023retrospective, turishcheva2024dynamic, turishcheva2025retrospective}, typically rely on deep architectures with numerous layers, contrasting sharply with biological visual processing where core computations are achieved through relatively few hierarchical stages. This architectural disparity suggests that biological systems implement more sophisticated computational primitives than the simple linear-nonlinear operations found in standard CNNs \citep{fitzgerald2015nonlinear, gilbert2007visual, zetzsche1990image, zetzsche1993importance}.

Natural scenes contain complex correlations and higher-order statistics that extend beyond simple linear relationships, including spatial dependencies and spatiotemporal interactions. Early visual systems are known to account for these features, across species from drosophila \citep{clark2014flies} \citep{fitzgerald2015nonlinear}, to frogs \cite{lettvin1959frog}, rabbits \cite{barlow1965mechanism}, cats \cite{hubel1965receptive} \cite{gilbert1977laminar}, macaques \cite{liu2021predictive}, and humans \cite{clark2014flies}. Importantly, these correlations exhibit a hierarchical structure, with higher-order correlations becoming increasingly sparse \citep{koenderink2018local}. Traditional CNNs, particularly those with limited depth, often struggle to effectively exploit these higher-order correlations \citep{azeglio2024convolution}.

We propose a novel approach that extends traditional 3D CNNs by incorporating higher-order operations within the convolutional operator itself, akin to a Volterra expansion \citep{volterra1887sopra, volterra1930theory}. This modification allows direct modeling of multiplicative interactions between neighboring pixels across time, enabling more effective capture of complex spatiotemporal dynamics even in shallow architectures.

We evaluate our approach on two distinct datasets: salamander retinal ganglion cell (RGC) responses to natural scenes \citep{maheswaranathan2023interpreting}, and a new dataset of mouse RGC responses to controlled geometric transformations that we collected using multi-electrode array recordings. Our contributions include:

1. A learnable higher-order convolution operator for video processing that captures complex spatiotemporal correlations more effectively than standard CNNs.

2. Demonstration of improved neural response prediction across different species and stimulus conditions, particularly with limited training data.

3. Evidence that our architecture naturally learns to encode fundamental geometric transformations, linking network performance to specific cell-type responses \citep{goetz2022unified, munch2009approach, kim2020dendritic, wang2021off, nirenberg1997light}.

Our results show that incorporating higher-order operations in early processing stages leads to significant improvements in neural response prediction, without increasing architectural depth. Notably, our model achieves superior performance with half of the training data compared to standard architectures \citep{mcintosh2016deep}, suggesting it better captures the underlying computational principles of biological vision.

\section{Biological Inspiration \& Related Work}
\label{BioAndRel}

Our work draws inspiration from biological visual systems, which implement sophisticated computations from earliest processing stages \citep{gollisch2010eye, kerschensteiner2022feature}. In the retina, cells exhibit responses to visual stimuli that cannot be explained by simple pointwise nonlinearities \citep{lettvin1959frog}, suggesting more complex computational mechanisms. 
The prevalence of such operations is particularly evident in motion detection and feature integration. For instance, the visual cortex contains specialized neurons that respond to complex features like corners and line endings \citep{hubel1965receptive}, performing advanced computations that integrate information across their receptive fields. These biological systems implement sophisticated operations that consider spatial relationships between multiple input points \citep{krieger1997higher, zetzsche1990image}, enabling effective extraction of spatiotemporal information. For spatiotemporal motion detection, direction-selective neurons in the fly visual system implement multiplicative interactions between spatially offset inputs with temporally delayed signals \citep{borst2015common}, while mammalian retinal ganglion cells achieve similar computations through asymmetric inhibitory synapses \citep{vaney2012direction}.

Recent research has highlighted fundamental limitations of artificial networks that rely solely on pointwise nonlinearities. Even single dendrites can solve problems like XOR \citep{xu2012nonlinear, ran2020type}, which pointwise operations struggle to model \citep{minsky1969introduction}. This suggests that incorporating non-pointwise operations could better align artificial systems with biological computation. Relatedly, in computational research, several approaches have explored nonlinear receptive fields. Early work analyzed inhomogeneous quadratic forms \citep{berkes2006analysis}, while recent developments have extended to image classification \citep{zoumpourlis2017non, azeglio2024convolution} and video processing \citep{roheda2024volterra}. Related approaches include bilinear CNN models \citep{lin2015bilinear} and compact bilinear pooling \citep{gao2016compact}, with applications in computational neuroscience \citep{ahrens2008nonlinearities}.

Our approach differs fundamentally by embedding higher-order operations within the convolutional operator itself, rather than applying them after feature extraction. This design enables the network to learn and apply complex nonlinear transformations throughout its entire depth, creating a more flexible architecture that better aligns with biological visual processing. By integrating these operations directly into the network's core building blocks, we achieve more efficient feature learning without increasing architectural depth.

\section{Beyond Pointwise Nonlinearities}
\label{beyondpointwise}

Traditional convolutional neural networks used in neural response prediction typically employ pointwise nonlinearities applied to weighted sums of inputs. When these nonlinearities are expanded as a polynomial series, a fundamental limitation becomes apparent: different orders of the expansion share the same weights, creating coupled terms across orders. Even for two pixels expanded at the second order, the six terms are characterised by only five parameters, therefore decreasing flexibility:

\begin{equation}
\label{tiedweightequation}
\begin{aligned}
& \sigma \left(\sum_{i = 1}^2 w_i x_i + b\right) \approx \\
& \alpha_0 + \alpha_1(w_1x_1 + w_2x_2) + \alpha_2(w_1x_1 + w_2x_2)^2 + \ldots \\
& = \alpha_0 + \color{red}{\alpha_1}\color{blue}{w_1}\color{black}{x_1} + \color{red}{\alpha_1}\color{green}{w_2}\color{black}{x_2} + \\
& \quad \color{orange}{\alpha_2}\color{blue}{w_1}\color{black}{^2x_1^2} + \color{orange}{\alpha_2}\color{green}{w_2}\color{black}{^2x_2^2} + 2\color{orange}{\alpha_2}\color{blue}{w_1}\color{green}{w_2}\color{black}{x_1x_2} + \ldots
\end{aligned}
\end{equation}

A quantitative analysis of these tied weights and their limitations was recently proposed by \citet{azeglio2024convolution}. While deeper or wider networks might theoretically compensate for this limitation \citep{bahri2020statistical}, such architectures diverge from the biological reality of neural processing, which achieves sophisticated computations with relatively few layers.
Building upon this analysis, we propose an extension to video processing that explicitly models multiplicative interactions between inputs across both space and time. Our approach, detailed in \textit{Section} \textbf{3D Higher-Order Convolution}, implements a learnable spatiotemporal receptive field model that captures the kinds of multiplicative interactions observed in biological visual processing \citep{zetzsche1990image}. 

\section{3D Higher-Order Convolution}
\label{Model}
To perform more complex computations while preserving the benefits of locality and weight sharing, we propose higher-order convolution for video processing. This approach extends the traditional convolution operator to include higher-order terms, enabling more sophisticated feature extraction that can better capture the complex dynamics of neural responses to visual stimuli.

We begin by considering an input spatiotemporal patch with dimensions along space (x, y) and time (t), reshaped as a vector \textbf{x}. Standard convolution can be expressed in terms of linear filtering, relating input \textbf{x} and output \textbf{y} as: $y(\textbf{x}) = b + \sum_{t = 1}^{T} \sum_{i = 1}^{H} \sum_{j = 1}^{W} w_1^{tij} x_{tij}$
where $w^{tij}_1$ represents the weights of the convolution kernel. To capture multiplicative interactions across space and time, we expand this function to include quadratic terms and beyond:

\begin{align}
y(\textbf{x}) = b &+ \sum_{t = 1}^{T} \sum_{i = 1}^{H} \sum_{j = 1}^{W} w_1^{tij} x_{tij} \nonumber \\
&+ \sum_{t,t'=1}^{T} \sum_{i,i'=1}^{H} \sum_{j,j'=1}^{W} w_2^{tij,t'i'j'} x_{tij} x_{t'i'j'} + ...
\end{align}

\begin{figure}[ht]
\centering
\includegraphics[width=0.45\textwidth]{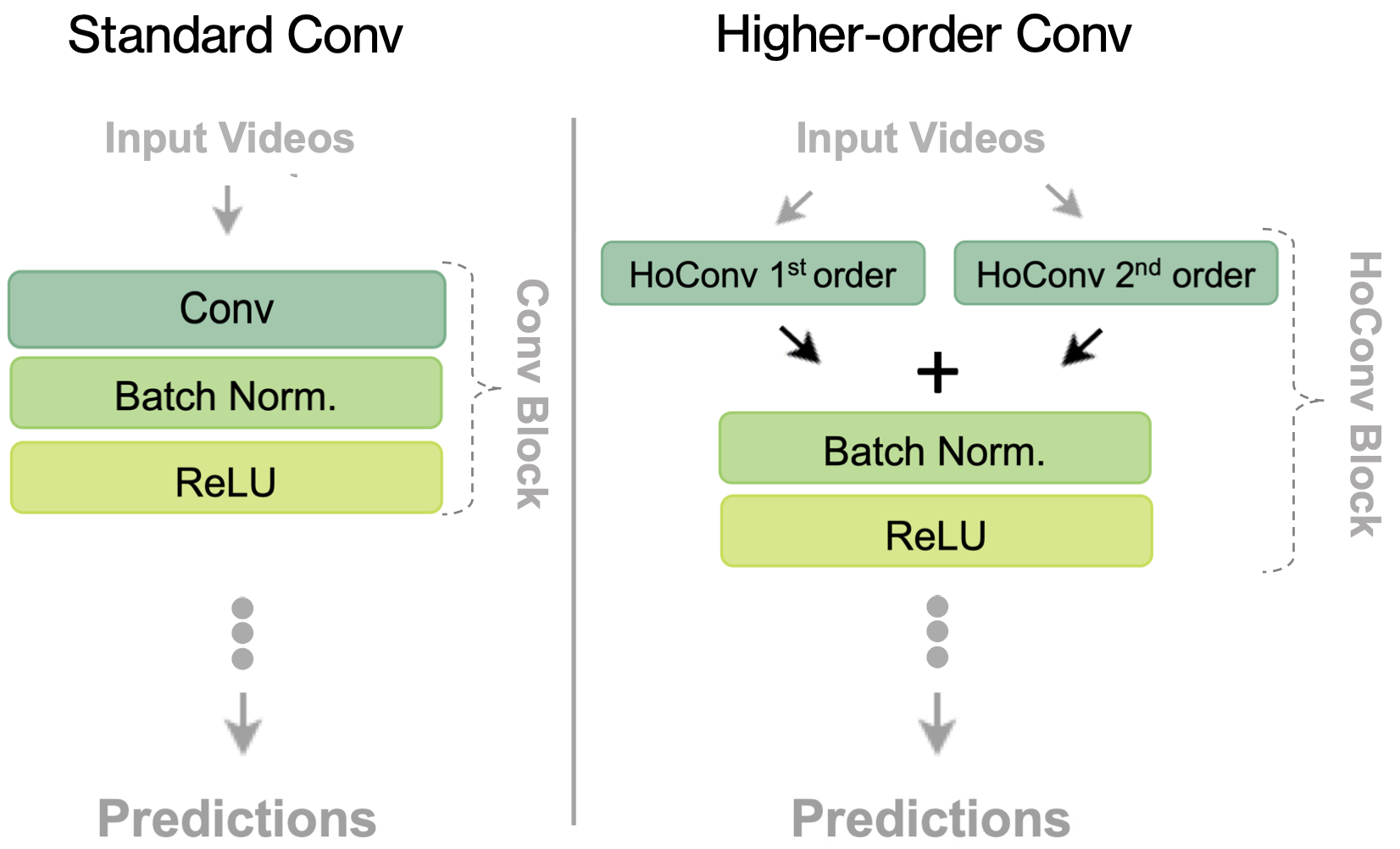}
\caption{Comparison of standard convolution (Conv) and higher-order convolution (HoConv) blocks.}
\label{Fig:architecture}
\end{figure}

For the entire video sequence, the higher-order convolution operation becomes:

\begin{align}
Y_{t,l,m} = b &+ \sum_{r,i,j} w_1^{rij} X_{t+r, l+i, m+j} \nonumber \\
&+ \sum_{r,i,j} \sum_{s,k,h} w_2^{rijskh} X_{t+r, l+i, m+j}X_{t+s, l+k, m+h} + ...
\end{align}

where $Y_{t,l,m}$ is the output at temporal position $t$ and spatial position $(l,m)$, $X$ is the input video, and $w_1$, $w_2$ are the learnable weights for first and second-order terms respectively. The sums over spatial dimensions $i$, $j$, $k$, $h$ and temporal dimensions $r$, $s$ run over the spatio-temporal convolution kernel dimensions for each order. The total number of parameters in this expansion grows as:
$n_V=(n_t n_s+p)! / (n_t n_s)! / p!$ where $p$ is the order of the term, $n_t$ is the temporal kernel size, and $n_s$ is the spatial kernel size.
To maintain balanced contributions across orders, we apply a scaling factor $s = \frac{1}{\sqrt{n_V}}$ to kernels of order greater than one, following \citet{zoumpourlis2017non} and \citet{azeglio2024convolution}. This scaling prevents higher-order terms from dominating network behavior during training while preserving their computational benefits. This spatio-temporal higher-order convolutional layer can functionally replace standard 3D convolutional layers in neural network architectures, as illustrated in Figure \ref{Fig:architecture}.

\section{Neural Response Prediction}

\begin{figure*}[t]
\centering
\includegraphics[width=0.8\textwidth]{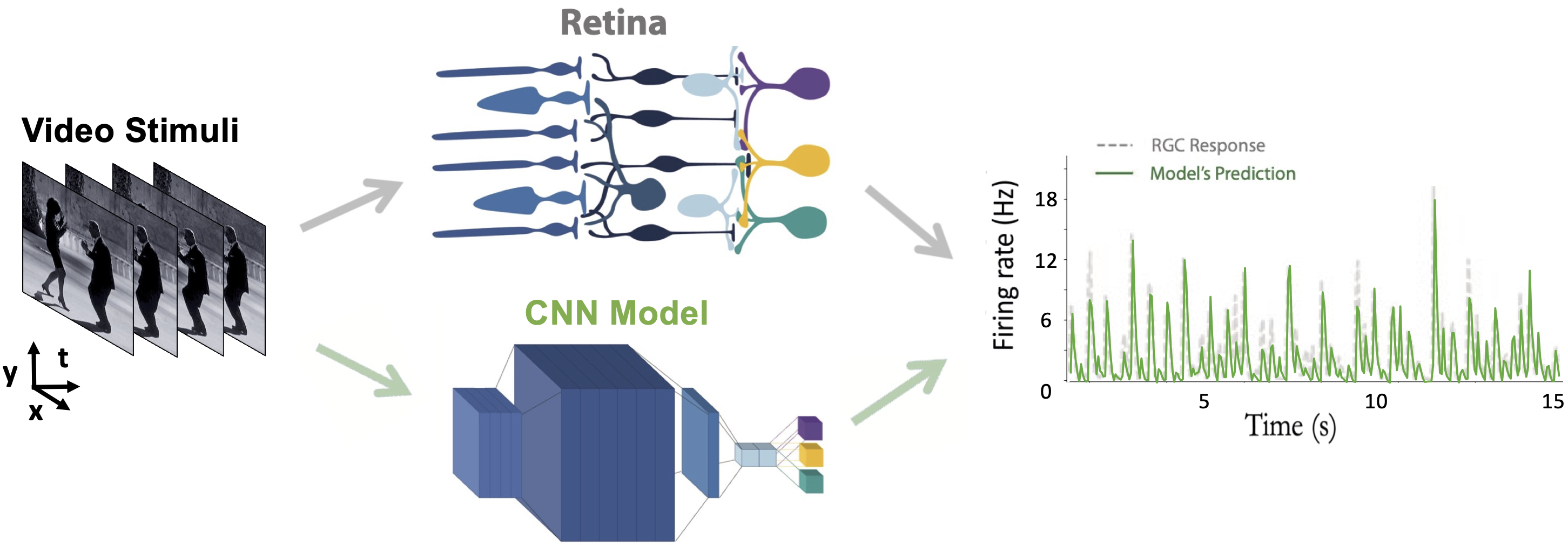}
\caption{Neural response prediction framework. Electrophysiological RGC responses to visual stimuli are first recorded using multi-electrode arrays. These recordings are then used to train convolutional neural networks to predict (multicell) neural responses from the visual input.} 
\label{Fig:framework}
\end{figure*}

We follow an established framework for studying neural computations where artificial neural networks are trained to predict RGC responses to visual stimuli \citep{mcintosh2016deep, klindt2017neural}. In this approach (see \textit{Figure} \ref{Fig:framework}), visual responses are first recorded from biological retinas using multi-electrode arrays during stimulus presentation. These recordings are then used to train artificial neural networks to predict the neural responses given the same visual inputs, allowing us to develop and test computational models of retinal processing.

To evaluate our higher-order convolution model within this framework, we first analyzed a public dataset from \citet{maheswaranathan2023interpreting}, comprising responses from \textbf{26 salamander RGCs} recorded across 3 experiments, with 5 trials in the test set. The visual stimuli consist of artificial videos generated by applying fixational eye movement-like jittering to natural images from the van Hateren database \citep{van1998independent}. While these stimuli capture spatio-temporal properties, they may not fully represent certain realistic correlations, such as those induced by approaching or receding objects. To address this limitation, we developed additional spatiotemporal stimuli by applying random perspective transformations to a binary checkerboard pattern, encompassing all possible geometric transformations of a 2D plane in 3D space (an anonymized sample of the stimulus is available at \href{https://streamable.com/pvrrob}{https://streamable.com/pvrrob}). This new dataset, recorded from mouse RGCs (we selected \textbf{40 cells}, see Section \textbf{Mouse RGCs Data} for details) using multi-electrode arrays, includes a training set of unique transformations and a test set of repeated transformations, totaling 60 repetitions in a single experiment.

\subsection{Salamander RGCs Data}

\begin{figure*}[t]
\centering
\includegraphics[width=\textwidth,height=\textheight,keepaspectratio]{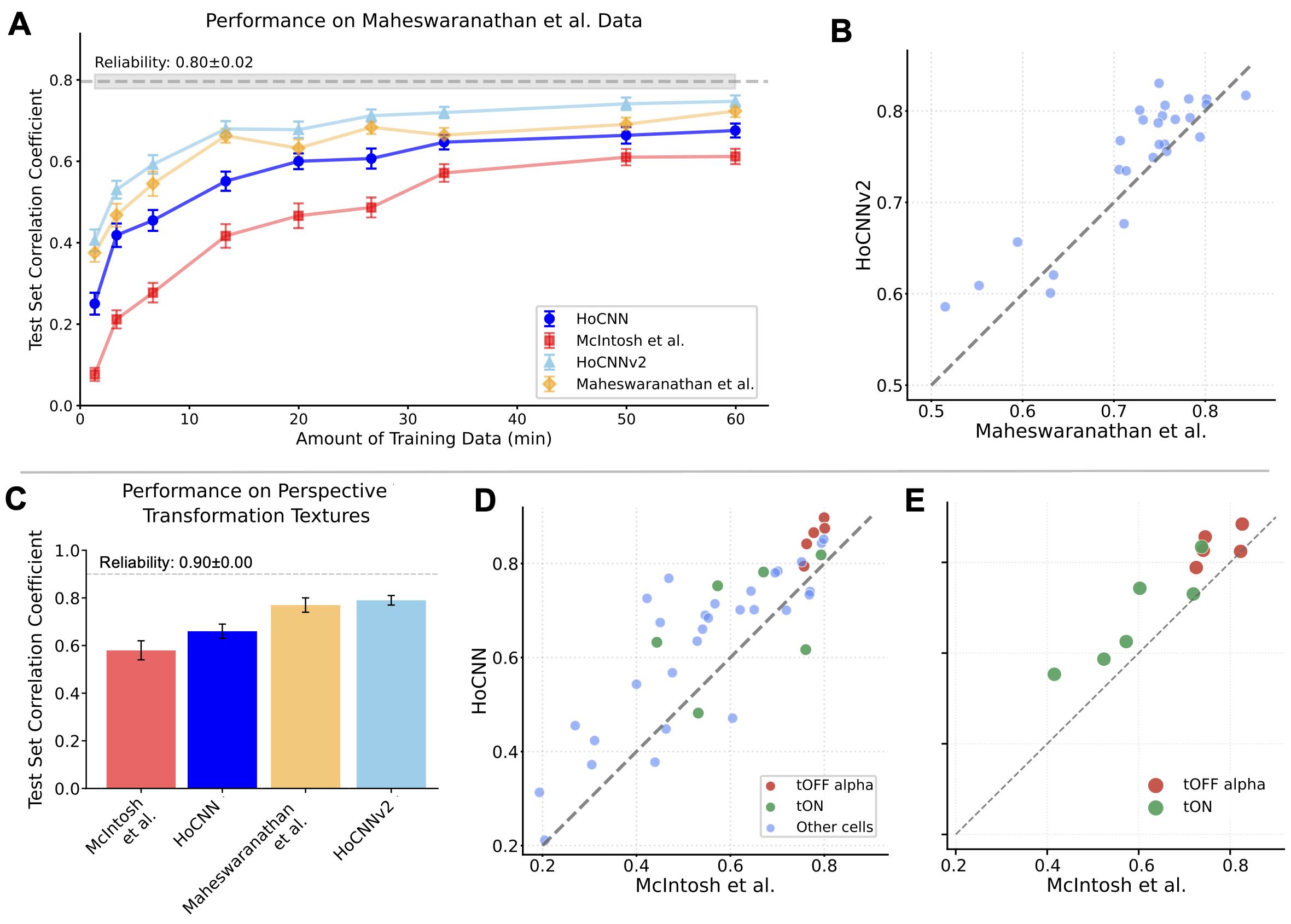}
\caption{\textbf{A.} Performance comparison across models and training set sizes for \citet{maheswaranathan2023interpreting} data. \textbf{B.} Single-cell correlation coefficients of the two best performing models: HoCNNv2 vs \citet{maheswaranathan2023interpreting} on held-out test data. \textbf{C.} Performance comparison across models for maximum size of training data (90\% vs 10\% validation). \textbf{D.} Single-cell correlation coefficients of the two considered models: HoCNN vs \citet{mcintosh2016deep} on held-out test data. \textbf{E.} Single-cell correlation coefficients HoCNN vs \citet{mcintosh2016deep}, trained on the tOFF-$\alpha$ + tON subset, on held-out test data. }
\label{Fig:results}
\end{figure*}

We first validated our approach by replicating results from \citet{mcintosh2016deep} and \citet{maheswaranathan2023interpreting}, confirming comparable performance on their datasets (see  \textit{Figure} \ref{Fig:results} A). We then first enhanced \citet{mcintosh2016deep}'s model by incorporating Higher-Order convolution, with the architectural differences illustrated in \textit{Figure} \ref{Fig:architecture}. We refer to this enhanced model as HoCNN.

Our primary modification involves extending the first convolutional layer to compute higher-order correlations. This specific placement follows findings from \citet{azeglio2024convolution}, which demonstrated optimal performance when higher-order computations directly interface with the superficial structure of natural scenes. If higher-order computations are delayed until later layers, earlier layers of the network may prematurely discard higher-order correlations, making it impossible for later layers to recover those correlations effectively. 

Comparative analysis between the original \citet{mcintosh2016deep} model and our HoCNN reveals a $11.5\%$ performance advantage for our approach (\textit{Figure} \ref{Fig:results} A and \textit{Table} \ref{table:performance_comparison_ganguli}). 
Notably, our HoCNN achieves equivalent performance to the original model with only half the training data, highlighting the importance of higher-order features in neural response prediction. 

Building on these promising results, we next investigated whether HoConv could enhance the performance of more recent architectures. Specifically, we focused on the model by \citet{maheswaranathan2023interpreting}, which has demonstrated superior performance for these salamander recordings. By incorporating HoConv in the first convolutional layer of their architecture, we created HoCNNv2. This modification yielded consistent improvements ($4.2\%$, see \textit{Figure} \ref{Fig:results} A and \textit{Table} \ref{table:performance_comparison_ganguli}). Our analysis of RGCs correlation coefficients on held-out test data, comparing the two best-performing models on \citet{maheswaranathan2023interpreting}'s dataset (\textit{Figure} \ref{Fig:results} B and \textit{Figure} \ref{AppFig:results_cell_by_cell} in \textbf{Appendix}), reveals systematic improvements. These enhancements bring our model's performance close to the theoretical maximum of retinal reliability (0.80 $\pm$ 0.02), consistent with the one found in \citep{mcintosh2016deep}. Given the limited amount of repetitions (5 trials), retinal reliability has been estimated with a leave-one out bootstrap approach.

\textbf{Training Details}. The base architecture from \citet{mcintosh2016deep} consists of 3 dendritic layers: 2 convolutional blocks (conv + batch norm + relu) and a fully connected layer with a softplus nonlinearity. 
HoCNN replaces only the first layer with HoConv \textit{Figure} \ref{Fig:architecture}). 
Both models were trained by minimizing a Poisson non-negative log-likelihood loss, using Adamw optimizer (lr = 5e-4, weight decay = 1e-6) with early stopping based on a 10\% validation split and ReduceLROnPlateau scheduler. For \citet{maheswaranathan2023interpreting}'s architecture, we similarly replaced the first convolutional layer with HoConv while maintaining all other architectural elements and training parameters as specified by the authors.

\begin{table*}[t]
\caption{Performance comparison (Correlation to Mean $\pm$ Standard Error to Mean) across different architectures and training data fractions (corresponding to data points in Figure \ref{Fig:results} \textbf{A} for \citet{maheswaranathan2023interpreting} data.}
\label{table:performance_comparison_ganguli}
\begin{center}
\small
\begin{tabular}{l|ccccccccc}
\multicolumn{1}{c|}{\bf Model} & \multicolumn{9}{c}{\bf Training Data Fraction (\%)} \\
 & 2 & 5 & 10 & 20 & 30 & 40 & 50 & 75 & 90 \\ \hline
McIntosh & 0.08±0.02 & 0.21±0.02 & 0.28±0.02 & 0.42±0.03 & 0.47±0.03 & 0.49±0.02 & 0.57±0.02 & 0.61±0.02 & 0.61±0.02 \\
HoCNN & 0.25±0.03 & 0.42±0.03 & 0.45±0.03 & 0.55±0.02 & 0.60±0.02 & 0.61±0.02 & 0.65±0.02 & 0.66±0.02 & 0.68±0.02 \\
Maheswaranathan & 0.38±0.02 & 0.47±0.03 & 0.55±0.03 & 0.66±0.02 & 0.63±0.02 & 0.68±0.02 & 0.66±0.02 & 0.69±0.02 & 0.72±0.02 \\
HoCNNv2 & \textbf{0.41±0.03} & \textbf{0.53±0.02} & \textbf{0.59±0.02} & \textbf{0.68±0.02} & \textbf{0.68±0.02} & \textbf{0.71±0.02} & \textbf{0.72±0.01} & \textbf{0.74±0.02} & \textbf{0.75±0.01} \\
\end{tabular}
\end{center}
\vspace{-15pt}
\end{table*}

\begin{table}[t]
\caption{Performance comparison (Correlation to Mean $\pm$ Standard Error to Mean) across different architectures and training data fractions (corresponding to data points in Figure \ref{Fig:results} \textbf{A} for \citet{maheswaranathan2023interpreting} data.}
\label{table:performance_comparison_our_data}
\begin{center}
\small
\begin{tabular}{l|c}
\multicolumn{1}{c|}{\textbf{Model}} & \textbf{Training Data Fraction (\%)} \\
& 90 \\
\hline
McIntosh et al. & 0.58$\pm$0.05 \\
HoCNN & 0.66$\pm$0.04 \\
Maheswaranathan et al. & 0.76$\pm$0.03 \\
HoCNNv2 & \textbf{0.79$\pm$0.02} \\
\end{tabular}
\end{center}
\vspace{-15pt}
\end{table}

\subsection{Mouse RGCs Data}

For our mouse RGCs responses, we selected the 40 most reliable cells based on test set performance. Cell reliability was computed using a bootstrap approach across repetitions. Specifically, for each bootstrap iteration, we randomly divided the trials into two groups, computed the mean responses for each group, and calculated the correlation between these means. This process was repeated 10,000 times to obtain robust reliability estimates with confidence intervals for each cell. This method provides a more stable estimate of reliability compared to the simple odd-even trial split approach, as it accounts for trial-to-trial variability more comprehensively.

We compared again \citet{mcintosh2016deep}'s, HoCNN, \citet{maheswaranathan2023interpreting}'s and HoCNNv2 models. Consistent with our results on salamander RGCs, HoCNN demonstrates $13.8\%$ improvement over \citet{mcintosh2016deep}'s baseline model (see \textit{Figure} \ref{Fig:results} C and Table \ref{table:performance_comparison_our_data}). This improvement is evident both across different training data fractions (see Appendix, Table \ref{tableApp:performance_comparison_perspective} and Figure \ref{AppFig:ourresults time}) and in cell-by-cell comparisons of the best-performing models (\textit{Figure} \ref{Fig:results} D). Relatedly, HoCNNv2 shows a $4.0\%$ improvement over \citet{maheswaranathan2023interpreting}'s model (see \textit{Figure} \ref{Fig:results} C and Table \ref{table:performance_comparison_our_data})

\textbf{Training Details.} Both models maintain the same architecture as described in the salamander experiments, with identical hyperparameters (Poisson non-negative log-likelihood loss, AdamW optimizer, lr = 5e-4, weight decay = 1e-6). Early stopping was implemented using 10\% of the training data as a validation set, along with a ReduceLROnPlateau scheduler. 

\section{Geometric Interpretation of HoCNN Performances}

\begin{figure}[t] 
\centering
\includegraphics[width=0.45\textwidth]{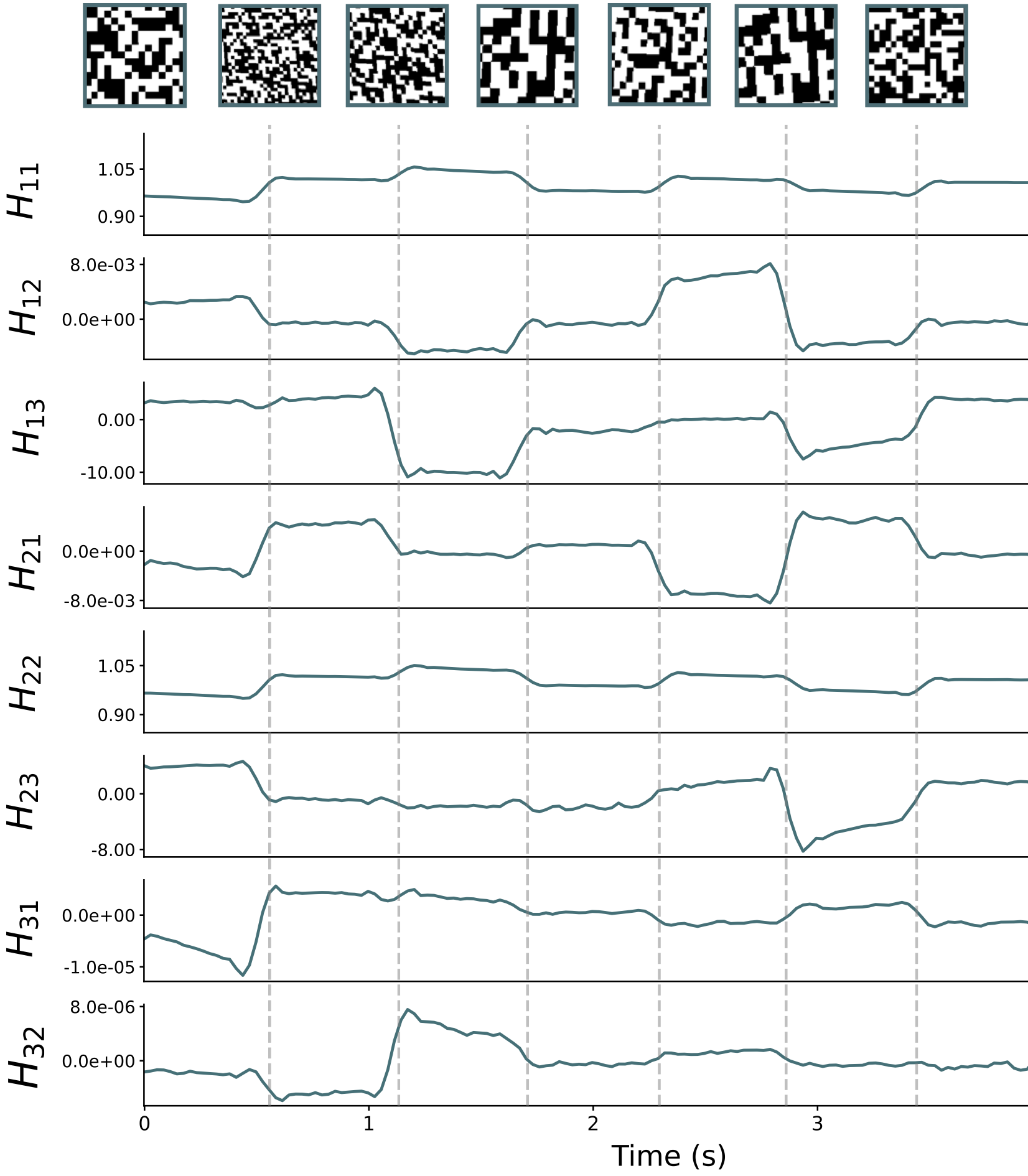}
\caption{Sample frames from our test set demonstrating perspective transformations applied to a checkerboard pattern. The transformation parameters are fully defined by the homography matrix $\mathbf{H}$. An anonymized sample of the stimulus is available at \href{https://streamable.com/pvrrob}{https://streamable.com/pvrrob}.
} 
\label{Fig:perspectivetrans}
\end{figure}

To understand why HoCNN outperforms standard architectures, we can examine how it processes spatiotemporal correlations in the visual input that give rise to specific motion patterns and geometric transformations. Our perspective stimuli are particularly well-suited for this analysis as they can be mathematically formalized using homographies - projective transformations between two planes that encompass common visual transformations like scaling, rotation, and perspective changes.
A homography describes how points in an image move during geometric transformations \citep{hartley2003multiple, szeliski2022computer}. For any point (x,y) in the original image, the homography matrix $\mathbf{H}$ determines its new position (x',y') after transformation:

\[
s
\begin{bmatrix}
x^{'} \\
y^{'} \\
1
\end{bmatrix} = \mathbf{H}
\begin{bmatrix}
x \\
y \\
1
\end{bmatrix} =
\begin{bmatrix}
H_{11} & H_{12} & H_{13} \\
H_{21} & H_{22} & H_{23} \\
H_{31} & H_{32} & H_{33}
\end{bmatrix}
\begin{bmatrix}
x \\
y \\
1
\end{bmatrix} 
\]

where $(x,y)$ and $(x',y')$ represent corresponding points before and after transformation, and $s$ is a scale factor. 
The elements of matrix $\mathbf{H}$ correspond to specific geometric operations:
\begin{itemize}
    \item \textit{Scaling} ($H_{11}$, $H_{22}$): control expansion/contraction.
    \item \textit{Rotation} ($H_{12}$, $H_{21}$): control rotational motion.
    \item \textit{Translation} ($H_{13}$, $H_{23}$): control translational motion.
    \item \textit{Perspective transformation} ($H_{31}$, $H_{32}$, $H_{33}$): control 3D viewing effects.
\end{itemize}


For example, when an object appears to approach the viewer, this primarily affects the scaling elements $H_{11}$ and $H_{22}$. The matrix is typically normalized with $H_{33}=1$ as homographies are defined up to a scale factor, ensuring a unique representation while preserving all geometric properties of the transformation. Figure \ref{Fig:perspectivetrans} illustrates how these parameters change during different transformations in our test set.

\subsection{Extracting Geometric Transformation Parameters}

Our analysis reveals that HoCNN naturally learns to encode geometric transformations while solving the neural prediction task, with particularly strong representation of scaling parameters that characterize object expansion and contraction. This alignment between network representations and fundamental geometric transformations may explain its superior performance in neural response prediction. While both our model comparisons (HoCNN vs \citet{mcintosh2016deep} and HoCNNv2 vs \citet{maheswaranathan2023interpreting}) isolate the effect of higher-order operations in their first layer, we focused our geometric feature analysis on the simpler architectures (HoCNN and \citet{mcintosh2016deep}'s) as their shallow structure both reflects the biological reality of early visual processing and allows direct attribution of improvements to higher-order operations, avoiding the complexity of tracing geometric information through multiple processing stages. Using their best-performing versions with frozen weights, we extracted features from the second convolutional block output of both HoCNN and \citet{mcintosh2016deep}'s model and trained a linear regressor to predict the eight free parameters of the homography matrix \textbf{H}. The regressor was trained on the training set and evaluated on the test set, providing a direct measure of how well each network captured geometric transformations. We focused particularly on the scaling parameters $H_{11}$ and $H_{22}$ as these characterize object expansion and contraction in the visual field - transformations that create complex spatiotemporal correlations that our higher-order convolutional layer was specifically designed to capture. Following \citet{fitzgerald2015nonlinear}'s demonstration of the importance of 3-point correlations in capturing such transformations, our architectural choice should theoretically enable more direct modeling of these geometric transformations.

\begin{figure}[t]
\centering
\includegraphics[width=0.48\textwidth]{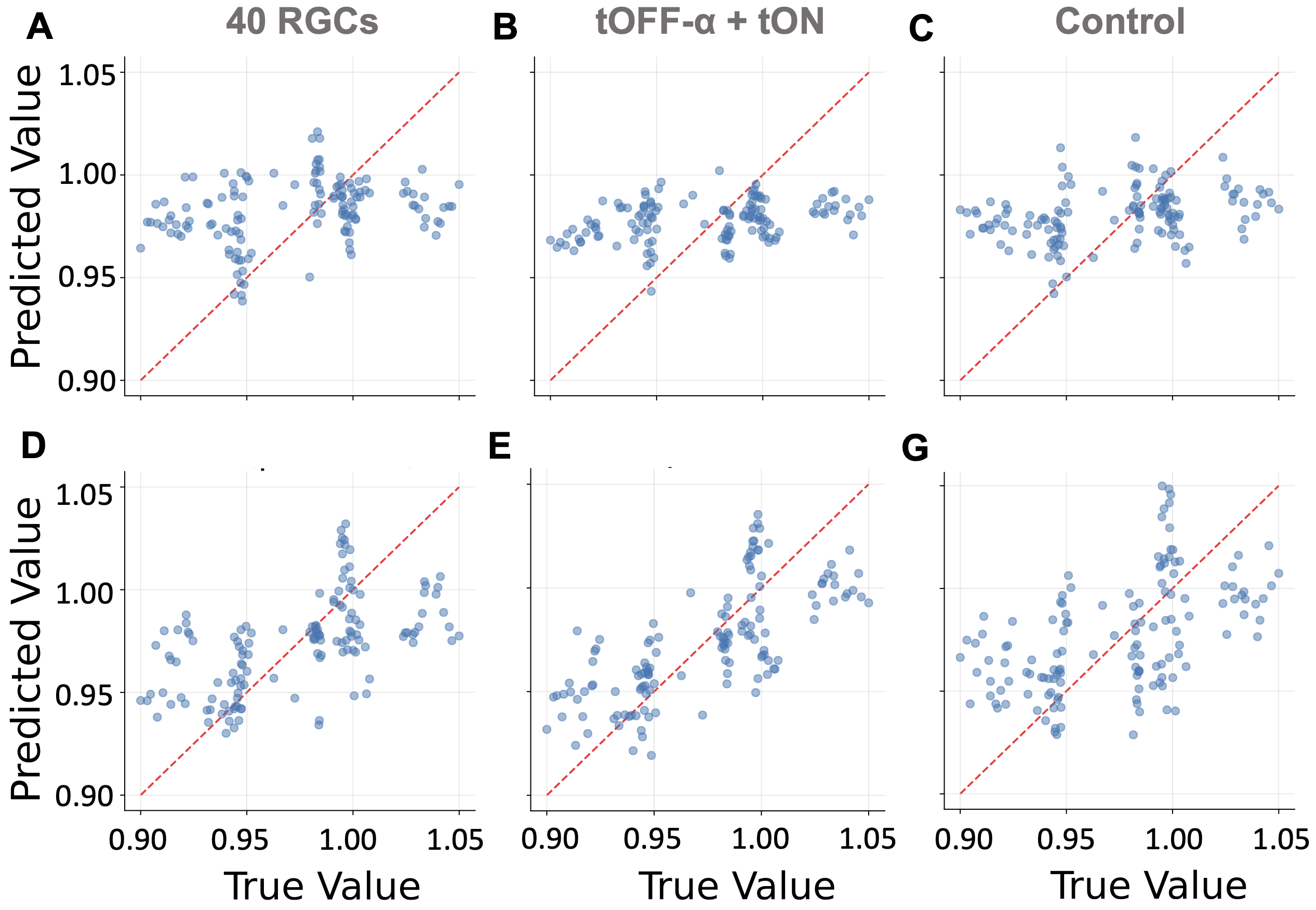}
\caption{Distributions of predicted values for scaling parameters $H_{11}$ from test data, comparing \citet{mcintosh2016deep}'s (\textbf{A}, \textbf{B}, \textbf{C}) and HoCNN model (\textbf{D}, \textbf{E}, \textbf{F}) against ground truth. \textbf{A} Depicts predicted values (linear regressor) against true value for scaling parameter $H_{11}$ from \citet{mcintosh2016deep}'s model (originally trained to predict all the 40 RGCs) features, correlation is $\rho_{McIntosh} = 0.29$; \textbf{B} Same values as \textbf{A}, but from features extracted from \citet{mcintosh2016deep}'s model trained to predict tOFF-$\alpha$ and tON (11 cells), correlation is $\rho_{McIntosh} = 0.32$; \textbf{C} Same values as \textbf{A} and \textbf{B}, but from features extracted from \citet{mcintosh2016deep}'s model trained to predict control RGCs subset (11 cells), correlation is $\rho_{McIntosh} = 0.32$. \textbf{D}, \textbf{E}, \textbf{F} depict the same results as \textbf{A}, \textbf{B}, \textbf{C} from HoCNN's features:  \textbf{D}'s $\rho_{HoCNN} = 0.56$; \textbf{E}'s $\rho_{HoCNN} = 0.72$ and \textbf{F}'s $\rho_{HoCNN} = 0.51$. } 
\label{Fig:h11}
\end{figure}

\begin{figure}[t]
\centering
\includegraphics[width=0.48\textwidth]{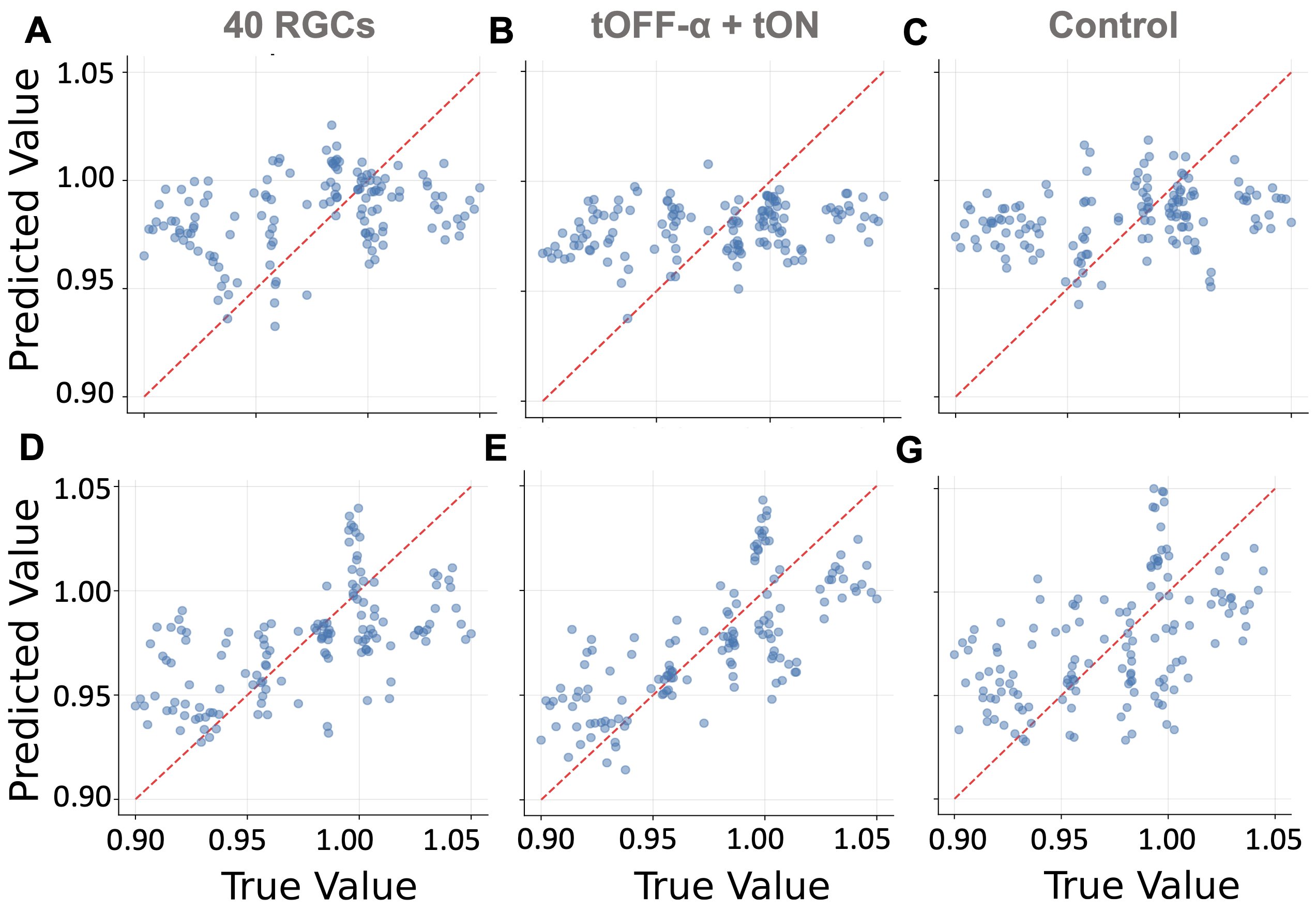}
\caption{Distributions of predicted values for scaling parameters $H_{22}$ from test data, comparing \citet{mcintosh2016deep}'s (\textbf{A}, \textbf{B}, \textbf{C}) and HoCNN model (\textbf{D}, \textbf{E}, \textbf{F}) against ground truth. \textbf{A} Depicts predicted values (linear regressor) against true value for scaling parameter $H_{22}$ from \citet{mcintosh2016deep}'s model (originally trained to predict all the 40 RGCs) features, correlation is $\rho_{McIntosh} = 0.33$; \textbf{B} Same values as \textbf{A}, but from features extracted from \citet{mcintosh2016deep}'s model trained to predict tOFF-$\alpha$ and tON (11 cells), correlation is $\rho_{McIntosh} = 0.28$; \textbf{C} Same values as \textbf{A} and \textbf{B}, but from features extracted from \citet{mcintosh2016deep}'s model trained to predict control RGCs subset (11 cells), correlation is $\rho_{McIntosh} = 0.28$. \textbf{D}, \textbf{E}, \textbf{F} depict the same results as \textbf{A}, \textbf{B}, \textbf{C} from HoCNN's features:  \textbf{D}'s $\rho_{HoCNN} = 0.57$; \textbf{E}'s $\rho_{HoCNN} = 0.73$ and \textbf{F}'s $\rho_{HoCNN} = 0.50$. } 
\label{Fig:h22}
\end{figure}

As shown in Tables \ref{table:scaling_correlations}, \ref{table:scaling_correlations_h22} and Figures \ref{Fig:h11}, \ref{Fig:h22}, HoCNN's features demonstrate substantially stronger correlations with the true scaling parameters. The correlation coefficient for $H_{11}$ is more than twice as high for HoCNN (0.56) compared to \citet{mcintosh2016deep}'s model (0.24). Similar considerations hold for $H_{22}$: $\rho_{HoCNN} = 0.57$ vs $\rho_{McIntosh} = 0.33$ . This improvement becomes even more pronounced when examining specific cell types known to process these transformations.

\begin{table}[t]
\caption{Correlation coefficients between predicted and true scaling parameters ($H_{11}$), comparing McIntosh and HoCNN performance for models trained on predicting different RGCs subsets: all the cells (40 cells), 11 cells (two families:  tOFF-$\alpha$ and tON) and another subset of 11 cells excluding tOFF-$\alpha$ and tON.}
\label{table:scaling_correlations}
\begin{center}
\small
\begin{tabular}{lcc}
\bf Predicted RGCs & \bf $\rho_{McIntosh}$ & \bf $\rho_{HoCNN}$ \\
\hline
40 (All Cells) & 0.29 & 0.56 \\
11 (\textbf{tOFF-$\alpha$ + tON}) & 0.32 & \textbf{0.72} \\
11 (\textbf{no} tOFF-$\alpha$ + tON) & 0.32 & 0.51
\end{tabular}
\end{center}
\vspace{-15pt}
\end{table}

\begin{table}[t]
\caption{Correlation coefficients between predicted and true scaling parameters ($H_{22}$), comparing McIntosh and HoCNN performance for models trained on predicting different RGCs subsets: all the cells (40 cells), 11 cells (two families:  tOFF-$\alpha$ and tON) and another subset of 11 cells excluding tOFF-$\alpha$ and tON.}
\label{table:scaling_correlations_h22}
\begin{center}
\small
\begin{tabular}{lcc}
\bf Predicted RGCs & \bf $\rho_{McIntosh}$ & \bf $\rho_{HoCNN}$ \\
\hline
40 (All Cells) & 0.33 & 0.57 \\
11 (\textbf{tOFF-$\alpha$ + tON}) & 0.28 & \textbf{0.73} \\
11 (\textbf{no} tOFF-$\alpha$ + tON) & 0.28 & 0.50
\end{tabular}
\end{center}
\vspace{-15pt}
\end{table}

\subsection{Cell Types}

The biological relevance of our results becomes clear when considering the specialized roles of different retinal ganglion cell (RGC) types. Among the more than 40 RGC types identified in mouse retina \citep{goetz2022unified}, transient OFF-alpha (tOFF-$\alpha$) cells are particularly relevant to our analysis, as they specifically detect looming objects \citep{munch2009approach, kim2020dendritic, wang2021off}. Similarly, transient ON (tON) cells show sensitivity to object motion \citep{nirenberg1997light}.

We identified 5 tOFF-$\alpha$ cells and 6 tON cells among our 40 recorded cells (see Section \textbf{Experimental Setup \& Methods}). To test the specificity of our model's geometric feature extraction, we re-trained both architectures from scratch on two distinct 11-cell subsets: one containing these identified motion-sensitive cells (average correlation to mean: 0.76 for HoCNN and 0.69 for \citet{mcintosh2016deep}'s model, see Figure \ref{Fig:results} \textbf{E}), and a control subset with comparable baseline performance (average correlation to mean: 0.78 for HoCNN and 0.70 for \citet{mcintosh2016deep}'s model). The results reveal a striking pattern: HoCNN's correlation with scaling parameters improves markedly ($\rho_{H_{11}}$ from 0.56 to 0.72; $\rho_{H_{22}}$ from 0.57 to 0.73) when trained on the motion-sensitive subset, while showing no such improvement ($\rho_{H_{11}} = $ 0.51; $\rho_{H_{22}} = $ 0.50) for the control group. McIntosh's model showed no such specialization, maintaining similar correlations ($\rho_{H_{11}}$ = 0.32; $\rho_{H_{22}}$ = 0.28) across both subsets.


\section{Discussion}

We have introduced a higher-order convolutional architecture for neural response prediction that demonstrates significant improvements over existing approaches without increasing architectural depth. Our results provide several key insights into both artificial and biological visual processing.
\textbf{First}, incorporating higher-order operations in early processing stages leads to more efficient learning, with our models achieving superior performance using substantially less training data (see also Figures \ref{AppFig:ourresults time} and \ref{AppFig:results_cell_by_cell} in \textbf{Appendix}). This suggests that explicitly modeling multiplicative interactions enables more effective capture of fundamental visual features, potentially mirroring computational strategies employed by biological visual systems.
\textbf{Second}, our analysis reveals that networks trained to predict neural responses naturally learn to encode geometric transformations of the visual input (see Tables \ref{table:scaling_correlations}, \ref{table:scaling_correlations_h22}; and Figures \ref{Fig:h11}, \ref{Fig:h22}). The superior performance of HoCNN in extracting these transformations, particularly scaling parameters, suggests that higher-order operations may be crucial for processing complex spatiotemporal patterns in natural scenes. 
\textbf{Third}, the improved performance on transient OFF-alpha cells, known to be critical for detecting looming objects, and transient ON cells, linked to object motion sensitivity, provides a concrete link between architectural choices and specific biological functions. This cell-type-specific improvement suggests that HoCNN's superior performance stems from its ability to better capture the geometric transformations these cells naturally encode. The higher-order operations in our model appear to provide a computational framework that aligns with the specialized processing capabilities of these biological motion detectors. While these results suggest a plausible mechanism linking geometric feature extraction to cell-type-specific performance, establishing causal relationships between these capabilities remains an important direction for future research. Further investigation could reveal how specific architectural choices in neural networks map to the processing of distinct visual features by different RGC types.

\subsection{Future Directions}
These findings open several promising research directions:

1. \textbf{Biological Mechanisms:} Investigation of how higher-order operations in artificial networks relate to biological computations, particularly gain control and adaptation mechanisms. The ability of higher-order convolutions to capture complex spatiotemporal correlations might provide a computational framework for understanding how neural circuits implement adaptive processing. This connection could be especially relevant for modeling contrast adaptation and dynamic range adjustment in early visual processing.

2. \textbf{Hierarchical Processing:} Extension of our approach to deeper visual processing stages, exploring how higher-order operations at different levels might capture increasingly abstract features.

3. \textbf{Geometric Transformations and Equivariance:} Further investigation of how neural circuits might implement transformation-aware processing, leveraging our finding that higher-order operations naturally encode geometric transformations. This framework could help explain how the visual system achieves stable object representations despite continuous transformations in the visual input, suggesting equivariance as a fundamental principle of neural computation.

More broadly, our results demonstrate that incorporating biologically-inspired computational primitives can lead to more efficient and interpretable visual processing systems. The success of our approach in both prediction accuracy and mechanistic interpretability suggests that bridging the gap between artificial and biological computation may require rethinking basic architectural elements rather than simply scaling existing designs.

\section{Experimental Setup \& Methods}

\subsection{Recordings}
Recordings were performed on isolated retinas from 1 C57BL6/J adult mice aged 5 months. The animals were housed in enriched cages with ad libitum food, and watering. The ambient temperature was between 22 and 25 \degree C, the humidity was between 50 and 70\% and the light cycle was 12--14 h of light, 10--12 h of darkness. The Animal was euthanized with CO2 4\% inhalation followed by cervical dislocation under dim red illumination according to institutional animal care standards of our University (will be disclosed upon acceptance). The retina was isolated from the eye under dim illumination and transferred as quickly as possible into oxygenated Ames medium (Merck, A1420). The retina was extracted from the eye cup and lowered with the ganglion cell side against a multi-electrode array whose electrodes were spaced by 30 \textmu m\textsuperscript{10}. During the recordings, the Ames' medium temperature was maintained at 37 \degree C. Raw voltage traces were digitized and stored for off-line analysis using a 252-channel preamplifier (MultiChannel Systems, Germany) at a sampling frequency of 20 kHz. The activity of single neurons was obtained using Spyking Circus V 1.1.0, a custom spike sorting software developed specifically for these arrays.

\subsection{Visual Stimulation}
Visual stimulations were presented using a white LED (MCWHLP1, Thorlabs Inc.) as light source and a Digital Mirror Device (DLP9500, Texas Instruments), focused on the photoreceptors using standard optics and an inverted microscope (Nikon). The light level corresponded to photopic vision: 4.9×104 and 1.4×105 isomerization / (photoreceptor. s) for S and M cones respectively.

The stimulus set included geometric transformation sequences of binary checkerboard patterns, comprising a training set of unique transformations (72 minutes) and a test set of repeated transformations with 60 trials per condition (5 minutes). We also presented chirp stimuli consisting of frequency-modulated full-field sine signals with varying temporal frequency and contrast levels, played at 50Hz with 20 repetitions of 32s length. Additionally, random binary checkerboard stimuli were used for receptive field mapping. 

All stimuli were presented with inter-stimulus intervals of 2-4s, maintaining background illumination at 10\textsuperscript{6} R* throughout the experiments.

\subsection{Receptive Field Mapping}
Receptive fields were mapped using reverse correlation analysis of responses to random binary checkerboard stimuli presented at 40 Hz for 40-60 minutes. For each cell, we computed a three-dimensional Spike-Triggered Average (STA) spanning two spatial dimensions and time (1s window, 40 time bins). The resulting spatiotemporal filters were decomposed into separate spatial and temporal components using Singular Value Decomposition. This analysis, provided precise localization of each cell's receptive field center and surrounding structure, essential for subsequent functional classification.

\subsection{Functional Cell Typing}
Cell classification was performed using a combination of:
1. Response profiles to chirp stimuli .
2. Spatiotemporal receptive field properties derived from STA analysis on random binary checkerboards.
3. Mosaic organization analysis based on receptive field centers.

Cells were initially clustered using a hierarchical unsupervised algorithm \citep{baden2016functional} applied to these response features. Cluster assignments were then manually verified based on established physiological criteria for mouse RGC types, with particular attention to identifying OFF-alpha transient and ON transient cells. This classification was further validated by analyzing the completeness and regularity of the resulting type-specific mosaics, a hallmark of RGC organization.

\section{Acknowledgments}
In compliance with the double-blind review process, author information, animal care standards, ethics guideline for experiments, and specific acknowledgments have been omitted from this submission. A comprehensive acknowledgment section, as well
as the code implementation, will be included upon acceptance of the paper.


\bibliographystyle{ccn_style}

\bibliography{ccn_style}

\clearpage
\section{Appendix}

\begin{figure}[h!]
\centering
\includegraphics[width=0.48\textwidth]{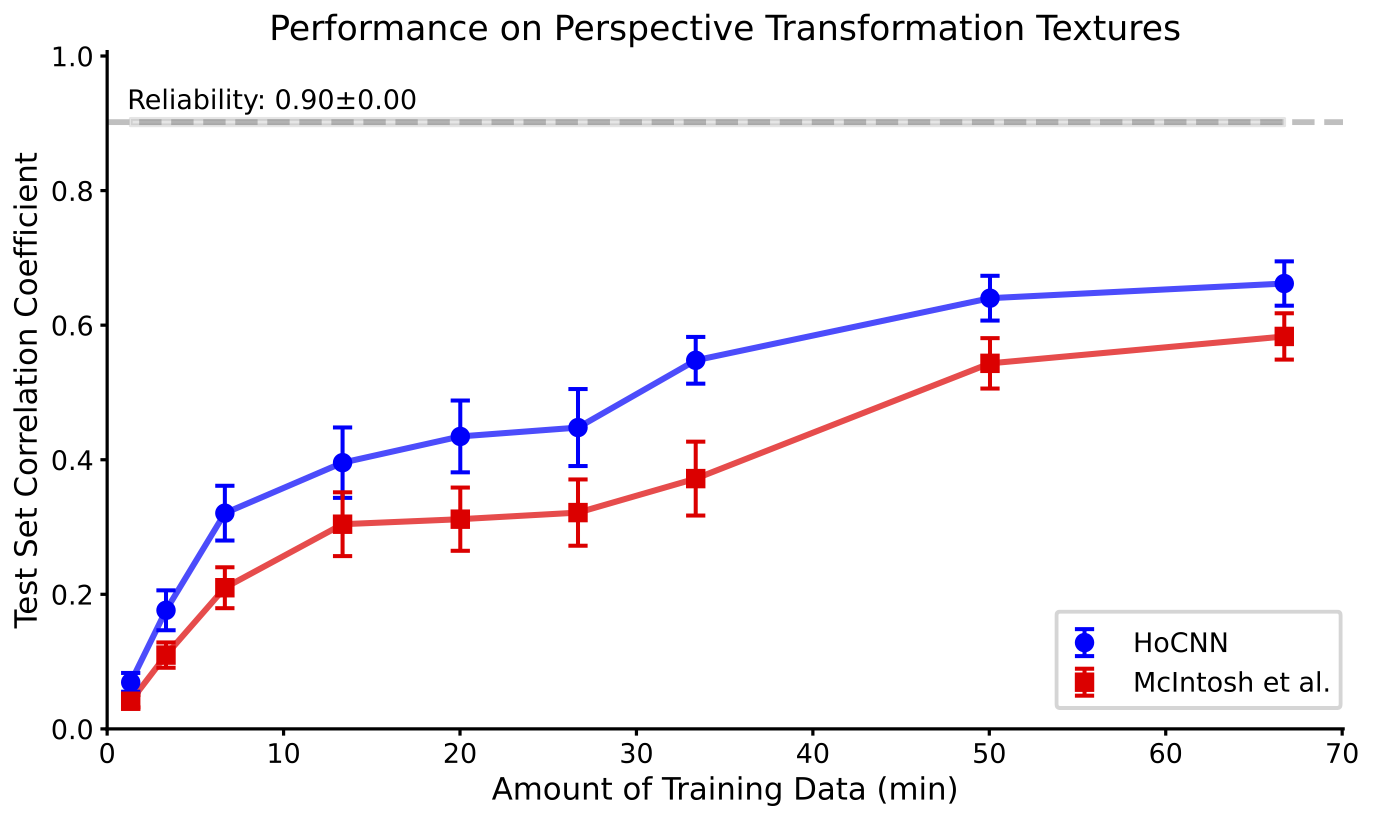}
\caption{Performance comparison across models - HoCNN and \citet{mcintosh2016deep} - and training set sizes for our data}
\label{AppFig:ourresults time}
\end{figure}

\begin{table*}[h!]
\caption{Performance comparison (Correlation to Mean $\pm$ Standard Error to Mean) across different architectures and training data fractions (corresponding to data points in \ref{Fig:results} \textbf{C} for neural responses to our perspective transformation textures.}
\label{tableApp:performance_comparison_perspective}
\begin{center}
\small
\begin{tabular}{l|ccccccccc}
\multicolumn{1}{c|}{\bf Model} & \multicolumn{9}{c}{\bf Training Data Fraction (\%)} \\
 & 2 & 5 & 10 & 20 & 30 & 40 & 50 & 75 & 90 \\ \hline
McIntosh & 0.04±0.01 & 0.11±0.03 & 0.21±0.04 & 0.30±0.05 & 0.31±0.05 & 0.32±0.06 & 0.37±0.03 & 0.54±0.03 & 0.58±0.03 \\
HoCNN & \textbf{0.07±0.01} & \textbf{0.18±0.02} & \textbf{0.32±0.03} & \textbf{0.40±0.05} & \textbf{0.43±0.05} & \textbf{0.45±0.05} & \textbf{0.55±0.05} & \textbf{0.64±0.04} & \textbf{0.66±0.03} \\
\end{tabular}
\end{center}
\vspace{-15pt}
\end{table*}

\begin{figure*}[t]
\centering
\includegraphics[width=\textwidth,height=\textheight,keepaspectratio]{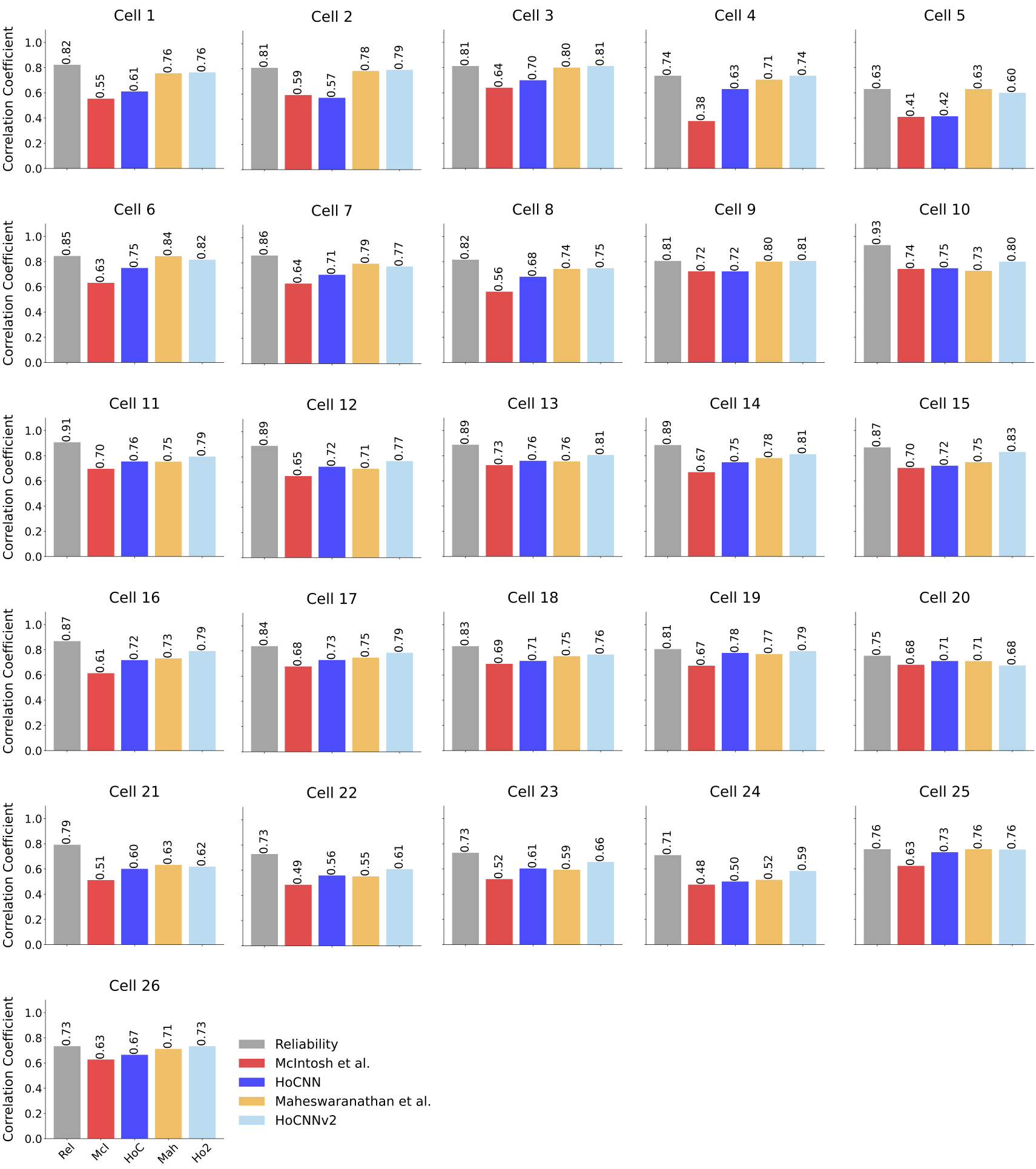}
\caption{Cell-wise performance comparison of \citet{mcintosh2016deep}'s, HoCNN, \citet{maheswaranathan2023interpreting}'s and HoCNNv2 models. on \citet{maheswaranathan2023interpreting}'s data}
\label{AppFig:results_cell_by_cell}
\end{figure*}

\end{document}